\def\year{2018}\relax
\documentclass[letterpaper]{article} 
\usepackage{aaai18}  
\usepackage{epsfig}
\usepackage{times}  
\usepackage{helvet}  
\usepackage{courier}  
\usepackage{url}  
\usepackage{graphicx}  
\usepackage{color}
\usepackage{mathtools}
\usepackage{amssymb}
\usepackage{amsmath}
\usepackage{textcomp}
\usepackage{bm}
\usepackage{algorithm2e}

\usepackage{booktabs}       
\usepackage{amsfonts}       
\usepackage{microtype}      
\usepackage{subfigure}
\usepackage{xcolor}
\usepackage{multirow}
\usepackage{longtable}
\DeclarePairedDelimiter{\ceil}{\lceil}{\rceil}
\newcommand{\specialcell}[2][c]{%
  \begin{tabular}[#1]{@{}c@{}}#2\end{tabular}}

\begin{document}
%
\title{Capturing Localized Image Artifacts through a CNN-based Hyper-image Representation}
\author{Parag Shridhar Chandakkar and Baoxin Li, Senior member, IEEE \\[1.5pt]
School of Computing, Informatics and Decision Systems Engineering \\
Arizona State University, Tempe, USA \\
\{pchandak, baoxin.li\}@asu.edu
}
\maketitle

\begin{abstract}
   Training deep CNNs to capture localized image artifacts on a relatively small dataset is a challenging task. With enough images at hand, one can hope that a deep CNN characterizes localized artifacts over the entire data and their effect on the output. However, on smaller datasets, such deep CNNs may overfit and shallow ones find it hard to capture local artifacts. Thus some image-based small-data applications first train their framework on a collection of patches (instead of the entire image) to better learn the representation of localized artifacts. Then the output is obtained by averaging the patch-level results. Such an approach ignores the spatial correlation among patches and how various patch locations affect the output. It also fails in cases where only few patches mainly contribute to the image label. To combat these scenarios, we develop the notion of hyper-image representations. Our CNN has two stages. The first stage is trained on patches. The second stage utilizes the last layer representation developed in the first stage to form a hyper-image, which is used to train the second stage. We show that this approach is able to develop a better mapping between the image and its output. We analyze additional properties of our approach and show its effectiveness on one synthetic and two real-world vision tasks - no-reference image quality estimation and image tampering detection - by its performance improvement over existing strong baselines.
\end{abstract}

\section{Introduction}

\begin{figure}[!t]
\centering     
\subfigure[]{
\includegraphics[width=0.4\textwidth, height=5cm]{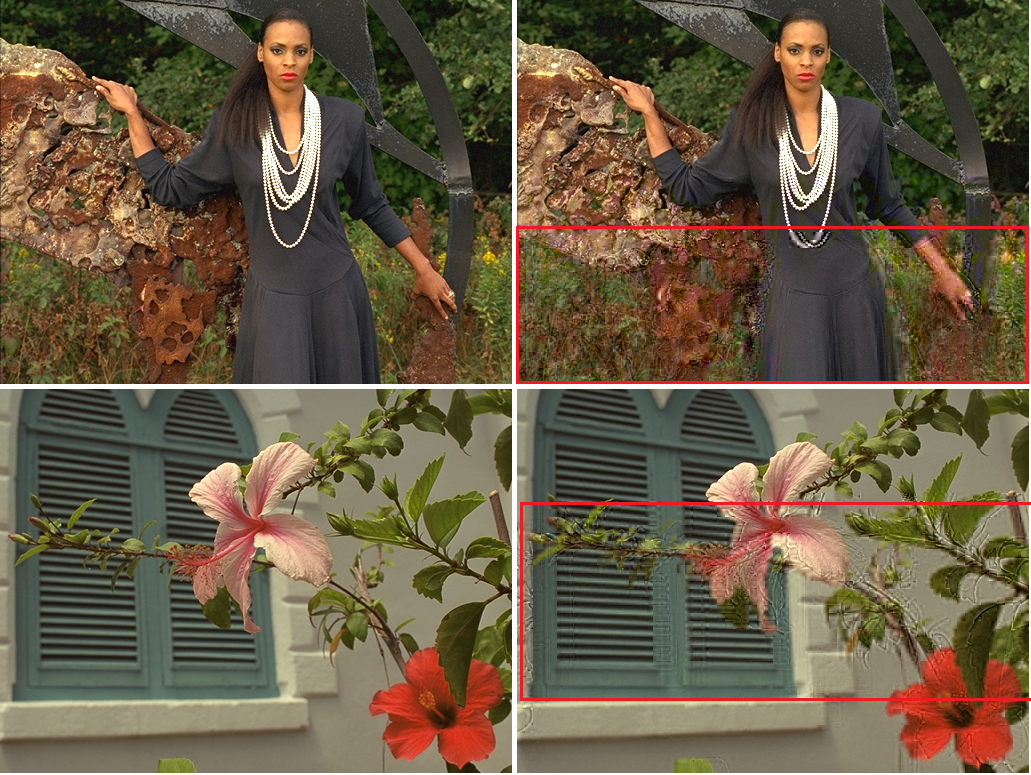}\label{fig:distortion_nonUniform} }
\subfigure[]{
\includegraphics[width=0.4\textwidth]{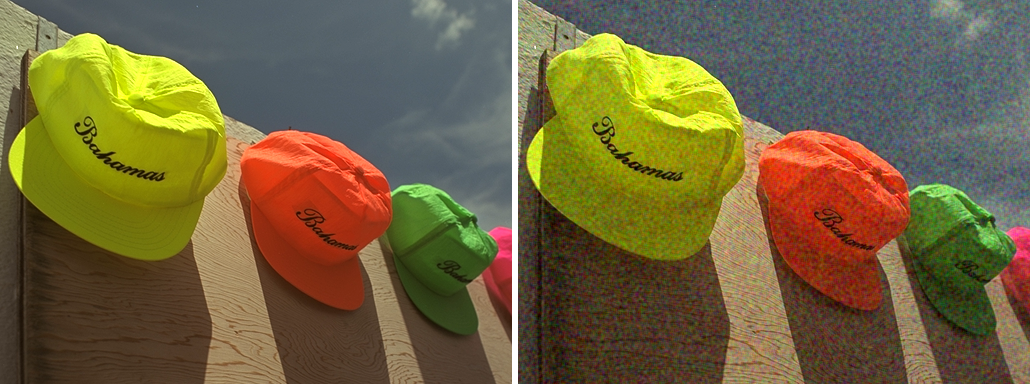}\label{fig:distortion_uniform}}
\subfigure[]{
\includegraphics[width=0.4\textwidth]{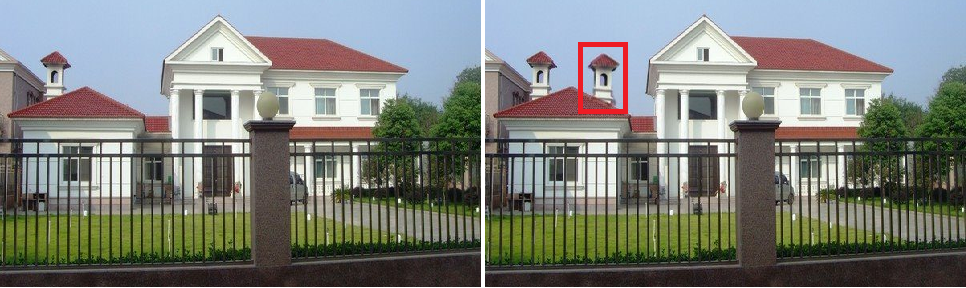}\label{fig:tampering}}
\caption[]{(a) and (b) Clean (left) and distorted (right) image pairs in the TID 2013 dataset. The images on the right in (a) and (b) are distorted by non-uniform and uniform noise respectively. (c) Authetic and forged image pair from CASIA-2 dataset. Red overlay shows distorted/forged regions in an image. Please zoom in for details and view the online version.}
\label{fig:problemIntro}
\end{figure}

Convolutional neural networks (CNN) have recently enjoyed tremendous success in many areas. In various fields, CNNs have surpassed the performance of previous state-of-art approaches by significant margins. For example, CNNs have shown considerable superiority in object recognition \cite{simonyan2014very,he2015deep}, face recognition \cite{taigman2014deepface}, semantic segmentation \cite{he2016semseg} etc. as well as in some understudied but long-standing and difficult problems such as gaze-following \cite{recasens2015they}. Since the advent of AlexNet  \cite{krizhevsky2012imagenet}, deep networks have also been continuously evolving for tasks such as segmentation \cite{zheng2015conditional}, object detection and image classification \cite{he2015deep,goodfellow2014generative,salimans2016improved}.

However, training CNNs to characterize localized image artifacts and label the image accordingly on a relatively small dataset remains a challenging task. With large amounts of data, deep CNNs may be able to learn a good representation for localized artifacts using a conventional pipeline (i.e. end-to-end training on images). Unfortunately, there are many applications where the labeled data is scarce, and the only way to obtain more data is by employing human experts, which is expensive and subject to their availability. This real-world constraint hinders the widespread use of advanced CNN architectures in such problems. On the other hand, the nature of some of these problems may exhibit properties that can be leveraged to increase the localization power as well as the volume of useful training data. For example, the images can be divided into smaller patches, and if the labels of these patches could be derived from the original image, 
 the number of labeled training samples could be increased potentially by a factor of, say 10-100, depending on how the patches are formed. Then CNNs could be trained on the augmented data, and image-level results could be derived by averaging patch-level results. Such patch-level training followed by averaging currently gives the state-of-art results for the problem of no-reference image quality estimation (NR-IQA) \cite{kang2014convolutional} (see Experiments section for details on NR-IQA).

The effectiveness of this patch-level training technique is only observed if one would assume that the artifacts are uniformly spread over the entire image, which is unfortunately too strong a constraint in practice. Certain real-world problems such as NR-IQA and image forgery classification are good examples where these strong constraints are often violated. Fig. \ref{fig:distortion_nonUniform} and \ref{fig:distortion_uniform} shows NR-IQA examples containing original (left) and distorted (right) images. The red overlay shows the distorted region in both the images. The distortions are localized and thus only few image patches are responsible for degrading its quality. Note that in the bottom image, the flower in the central salient region is distorted whereas in the upper image, some parts towards the lower non-salient region are distorted, preserving the quality of salient face. This affects the perceived quality as the top image (score = $5.56$) has been judged to be of higher quality than the bottom one (score = $3.53$) based on the extensive subjective tests conducted in \cite{ponomarenko2015image}. \textit{Interestingly, the type and the severity of the distortion added are identical for both images}. Thus the image quality is dependent on various factors such as distortion type, severity, affected patch locations and their texture etc. This cannot be handled by the aforementioned patch-level training technique. Similar observations can be made about Fig. \ref{fig:tampering}. The image needs to be categorized into authentic or forged. The forgery is localized (shown by the red bounding box) and may not be effectively captured by a conventional deep CNN-pipeline or independent patch-level training, especially when the training data is only in the order of thousands. Patch-level training is only effective in case of uniform distortion as shown in Fig. \ref{fig:distortion_uniform}.

To combat these scenarios, we present a novel CNN-based approach focused towards such type of data. Our approach does not require the image and its patches to share a similar distribution. It works well even if there is only one patch per image which plays an important role in determining the image label (relaxing the requirement that all patches from an image should equally contribute to the decision, e.g. by means of averaging patch-results). We evaluate our approach on one synthetic data and two real-world, challenging vision tasks - NR-IQA and image forgery classification. We will demonstrate that the proposed method produces a state-of-art performance for both the applications. We now explain the problem setup followed by our approach.

\section{Problem Setup} \label{sec:probSetup}

We consider a problem where our data has some localized information embedded in it that is crucial to getting the desired output. Additionally, the quantity of available data is limited and its artificial generation is non-trivial or even impossible. We assume that inferring important patches is possible or this information is provided as ground-truth.

Consider a database of images $\mathcal{X}$ containing $N$ images, and their labels, denoted by $(X_i, Y_i) \ \forall \ i$. An image has $m$ patches, $x_i^1,...,x_i^m$. Here, $x_i^j$ denotes $j^{\text{th}}$ patch in the $i^{\text{th}}$ image. We denote patch-level labels by $y_i^1,\hdots,y_i^m$. Patch-level labels can be inferred from the image or they can be a part of ground-truth labels. Thus training on patches and then averaging all the patch scores to obtain the image label is a naive way which, actually, works well in practice. However, we do not allow the learner to understand the relation between patch scores and the image score. In other words, the network cannot learn an optimal weighing strategy for all image regions, especially when only few regions contain the important localized artifacts. Training on the entire image with a deep stack of convolutional filters may achieve the desired result, but then limited amounts of data prevents CNN from generalizing well. Thus, our problem becomes, given an image $X_i$ and its patches $x_i^1,\hdots,x_i^m$, first infer patch-level labels - $y_i^1,\hdots,y_i^m$. Subsequently, develop a CNN framework that aggregates the information from the collection $({x_i^j},{y_i^j}) \ \forall i, \forall j$ and forms a mapping function $f: X_i \rightarrow Y_i, \ \ \forall \ i$.

\section{Proposed Approach} \label{sec:proposedApproach}

Our approach has two CNN stages out of which the first one is trained on a collection of labeled patches. Given a database of labeled images - $\mathcal{X}, \mathcal{Y}$ - we first extract all the patches and derive their corresponding labels. As mentioned earlier, the patch-level labels are either inferred from the image label or are provided as ground-truth, if available.

\subsection{Training the First Stage}

The first stage CNN follows a conventional training pipeline and is trained on a collection of labeled patches. We detail the pre-processing steps and the training procedure in Experiments section as they vary by application.

It is well-known and empirically verified that deeper layers encode increasingly powerful features \cite{zeiler2014visualizing,yosinski2015understanding} that are semantically more meaningful \cite{zhou2014object}. Thus after training the first stage, we extract the ReLU-activated responses of the last but one layer for all the patches in the train and validation set. During our experiments, we observed that ReLU-activated sparse response provides better validation loss than the pre-ReLU responses. This will be used as the input representation for the second stage described as follows.

\subsection{Training the Second Stage} \label{sec:TrainingSecondStage}

A trained first stage CNN provides a $D$-dimensional representation obtained from the last but one layer for any given image patch. Let us denote the $m$ overlapped patches for the $i^{\text{th}}$ image by $x_i^1, \hdots, x_i^m$. We extract the $D$-dimensional representations from these $m$ patches and arrange them in a $U \times V \times D$ hyper-image denoted by $H_i$. The arrangement is done such that the $(u,v)$ element of the hyper-image - $H_i^{uv}$ - corresponds to the representation of $(u,v)$ patch in the original image. In other words, each patch in the original image now corresponds to a point in $D$-dimensions in the hyper-image. Note that $U$ and $V$ are smaller as compared with the image height and width respectively, by a factor proportional to the patch size and overlap factor.

We now train a second stage CNN on the hyper-images and their labels, which shares architectural similarities with its counterpart - the first stage CNN. We do not perform mean-centering or any other pre-processing since these representations are obtained from normalized images.

Each pixel in the hyper-image being inputted to the second stage CNN is of the form $H_i^{uv} = f_1(f_2(\hdots (f_n(x_i^{uv})) \hdots)) \ \forall \ u, v$, where $f_1,\hdots,f_n$ represent the non-linear operations (i.e. max-pooling, convolutions, ReLUs etc.) on an image as it makes a forward pass through the first stage. Then in the second stage, the label is inferred as, $y_i = g_1(g_2(\hdots (g_n(H_i)) \hdots))$, where $H_i$ denotes the hyper-image being inputted to the second stage corresponding to the $i^{\text{th}}$ image and $g_1,\hdots,g_n$ denote the non-linear operations in the second stage. The following equation expresses $y_i$ as a highly nonlinear function of $x_i^{uv} \ \forall \ u, v$.

\begin{equation}
\begin{aligned}
y_i &= g_1(g_2(\hdots (g_n(\{H_i^{11}, \hdots, H_i^{uv}\})) \hdots)), \\ \text{where,} \ & H_i^{uv} = f_1(f_2(\hdots (f_n(x_i^{uv})) \hdots)) \ \forall \ u, v
\end{aligned}
\end{equation}

\noindent This allows the multi-stage CNN to take decisions based on context and by jointly considering all the image patches. Both the stages combined also provide us with higher representational capacity than had we trained it simply with patches followed by averaging. 

Note that the convolutional filters of the second stage CNN only operate on a small neighborhood at a time (usually $3 \times 3$), where each point in this neighborhood corresponds to the $D$-dimensional representation of a patch. So if the receptive field of a neuron in the first layer is $3 \times 3$, it is looking at 9 patches arranged in a square grid. Thus filters in the early layers learn simple relations between adjacent patches whereas the deeper ones will start pooling in patch statistics from all over the image to build a complex model which relates the image label and all patches in an image.

This two-stage training scheme raises an issue that it needs the first stage to produce a good representation of the data, since the second stage is solely dependent on it. Any errors in the initial stage may be propagated. However, in our experiments we find that the accuracy of the entire architecture is always more than the accuracy obtained by:

\begin{enumerate}
\item Training end-to-end with a deep CNN. \vspace{-6pt}
\item Training on patches and averaging patch scores over the entire image. \vspace{-6pt}
\item End-to-end fine-tuning of a pre-trained network.
\end{enumerate}

\noindent This points to two possibilities. Firstly, the second stage is powerful enough to handle slight irregularities in the first stage representation. This is expected since CNNs are resilient even in the presence of noise or with jittered/occluded images to some extent \cite{dodge2016understanding}. Secondly,   
inputting a hyper-image containing D-dimensional points to a CNN results in a  performance boost. For example, predicting a quality score for the distorted images shown in Fig. \ref{fig:distortion_nonUniform} and Fig. \ref{fig:distortion_uniform}  can be viewed as a regression task with multiple instances. A single patch having lower quality will not necessarily cause the entire image to have lower quality score. Similarly, an image with multiple mildly distorted patches at corners may have higher quality score than an image with a single severely distorted patch placed in a salient position. Thus quality score is related to the appearance/texture of all patches, their locations, distortion severity etc. Our network acquires location sensitivity to a certain extent (shown in experiments on synthetic data) as it learns on a $U \times V \times D$ grid. Distortion strengths of the individual patches are encoded in the D-dimensions, which are unified by the second stage to get desired output for a given image. On the other hand, an end-to-end conventional network pipeline will need to learn both - 1. the discriminative feature space and 2. the patches which contribute to the image label. The patch-averaging technique will fail as it will assign equal weight to all the individual patches potentially containing distortions of different strengths. To summarize, the proposed architecture attempts to learn the optimal mapping between image label and spatially correlated patches of that image.

\subsection{Testing}\label{sec:Testing}

Firstly, we extract a fixed number of patches from an image, extract their representations using the first stage and arrange it in a $U \times V \times D$ grid to form a hyper-image. Our approach does not require resizing of input images. Instead, we change the strides between patches at run-time to obtain the desired shape. In applications where resizing images could change the underlying meaning of images or introduce additional artifacts (e.g. image quality, image forgery classification), this approach could be useful. The procedure to compute the strides at run-time is as follows. we first compute (or assume) the maximum height and width among all the images in our dataset. If a test image exceeds those maximum dimensions, we scale it isotropically so that both its dimensions meet the size constraints. Lets denote the maximum height and width by $M$ and $N$ respectively. Thus the number of patches required in the $x$ and $y$ direction of the grid are $n_{p_x} = \ceil{\frac{N}{sz_x}}, n_{p_y} = \ceil{\frac{M}{sz_y}}$. Here, $sz_y \times sz_x$ is the patch size that is used in the first stage. For any new $\hat{M} \times \hat{N}$ image, the required strides (denoted by $s_x$ and $s_y$) to obtain a fixed number of patches in the grid are as follows:\vspace{-3pt}

\begin{equation} \label{eq:strides}
s_x = rnd\left(\frac{\hat{N}-sz_x}{n_{p_x}-1}\right), s_y = rnd\left(\frac{\hat{M}-sz_y}{n_{p_y}-1}\right)
\end{equation}

\noindent After obtaining the hyper-image $ \in \mathbb{R}^{U \times V \times D}$ from the first stage, we make a forward pass with it through the second stage to obtain the desired output. In the upcoming sections, we apply the proposed approach to a synthetic problem and two real-world challenging vision tasks, review relevant literature and discuss about other aspects of our approach.

\section{Experiments and Results} \label{sec:exptsAndResults}

We evaluate our approach on a synthetic task and two challenging real-word problems - 1. no-reference image quality assessment (NR-IQA) and 2. image forgery classification. Apart from the dependence on localized image artifacts, an additional common factor that aggravates the difficulty level of both these problems is that the amount of data available is scarce. Manual labeling is expensive as subject experts need to be appointed to evaluate the image quality or to detect forgery. Additionally, artificial generation of data samples is non-trivial in both these cases. We will now begin by describing the setup for our synthetic task.

\begin{table}[!b]
\centering
\caption{Results of experiments on synthetic data}
\label{table:results_synthetic}
\vspace{3pt}
\setlength{\tabcolsep}{0.68em}
\begin{tabular}{c c c}
\hline \hline \\[-9pt]
\multicolumn{3}{c}{\textit{Experiment 1 on synthetic data}} \\
\hline \\[-9pt]
Approach & SROCC & PLCC \\
\hline \\[-9pt]
Patch-averaging & $0.9132$ & $0.8982$ \\
\textbf{Proposed} & \bm{$0.9611$} & \bm{$0.9586$} \\
\end{tabular}
\begin{tabular}{c c}
\hline \hline \\[-9pt]
\multicolumn{2}{c}{\textit{Experiment 2 on synthetic data}} \\
\hline \\[-9pt]
Approach & Mean of SROCC and PLCC \\
\hline \\[-9pt]
Patch-averaging & $0.665,0.7,0.544,0.5,0.439$ \\
\textbf{Proposed} & \bm{$0.886$},\bm{$0.811$},\bm{$0.738$},\bm{$0.744$},\bm{$0.718$} \\
\hline \hline
\end{tabular}
\end{table}

\textbf{Synthetic task:} While this task is constructed to be simple, we have included certain features that will examine the effectiveness of the proposed approach. The task is to map an image to a real-valued score that quantifies the artifacts introduced in that image. The dataset contains $128 \times 128$ gray-scale images that have a constant colored background. The color is chosen randomly from $[0,1]$. We overlay between one to five patches on that image. Each patch contains only two shades - a dark one $\in [0,0.5)$ and a bright one $\in [0.5,1]$. A random percentage of pixels in a patch is made bright and the others are set to dark. Finally, all the pixels are scrambled. Size of each patch is $16 \times 16$. 

Let us denote the synthetic image by $S$ and the $i^\text{th}$ patch by $p_i$. The number of patches overlaid on $S$ is $\eta \ (\leq 5)$. Let $s_0$ denote the constant background value of $S$. $p_i^{jk}$ denotes the $(j,k)$ pixel of $p_i$. The center of the patch $p_i$ is denoted by $c_i$. The image center is denoted by $\hat{c}$. The score of this image can now be defined as $5 - \sqrt{\sum_{i=1}^{\eta} \alpha_i^2}$, where $\alpha_i$ is:\vspace{-3pt}

\begin{equation}
\alpha_i = \left(\sum_{j=1}^{16} \sum_{k=1}^{16} \left(\frac{|s_0 - p_i^{jk}|}{16 \times 16} \right)\right) + \left(1 - \frac{||c_i - \hat{c}||_2}{\text{dist}_N} \right)
\end{equation}

\noindent The first term computes the Manhattan distance between a background pixel and all the pixels of a patch. This can be viewed as a dissimilarity metric between the background and a patch. The second term imposes a penalty if the patch is too close to the image center. The term $\text{dist}_N$ is a normalization factor. If the patch lies at any of the four corners, then $ ||c_i - \hat{c}||_2 = \text{dist}_N$ and the penalty reduces to zero.  Higher score indicates presence of lower artifacts in an image.

We perform another experiment to test the sensitivity of our approach to the patch locations. To this end, we created $1$K images that all had identical background. The number of patches and their content (i.e. two shades and pixel scrambling) were also fixed across all $1$K images. The only variable was their positions in the image. we compared our two-stage approach and patch-averaging. Patch-averaging assigns nearly identical scores to all $1$K images whereas our approach gives higher correlation with the ground truth. When everything except patch positions was fixed, our approach had higher correlation and slow degradation with increasing number of patches. See Table \ref{table:results_synthetic} for results. Synthetic images are provided in supplementary.

\textbf{No-reference image quality assessment (NR-IQA): } Images may get distorted due to defects in acquisition devices, transmission-based errors etc. The task of IQA requires an automated method to assess the visual quality of an image. Conventional error metrics such as RMSE/PSNR cannot capture the correlation between image appearance and human-perception of the image quality. Two variants of this problem exist - full-reference IQA and no-reference IQA (NR-IQA). In the former task, an original image and its distorted counterpart is given. A quality score has to be assigned to the distorted one with respect to the original image. Few representative approaches that try to solve this problem are SSIM \cite{wang2004image}, MSSIM \cite{wang2003multiscale}, FSIM \cite{fsim}, VSI \cite{zhang2014vsi} etc. However, in real-world scenarios, we may not have a perfect, non-distorted image available for comparison. Thus NR-IQA variant has been proposed. In NR-IQA, a single image needs to be assigned a quality score with respect to a non-distorted, \textit{unobserved} version of that image. This score must correlate well with human perception of image quality. While creating ground-truth for this problem, a constant value is associated with a non-distorted image. This serves as a reference on the quality score scale. This problem involves developing a discriminative feature space to different kinds and degrees of distortions. Such setting is more suitable for learning schemes, which is reflected by the fact that most approaches used to tackle this problem belong to the learning paradigm. Few of the representative approaches include BRISQUE \cite{mittal2012no}, CORNIA \cite{ye2012unsupervised,ye2013real}, DIIVINE \cite{moorthy2011blind}, BLIINDS \cite{saad2012blind}, CBIQ \cite{mittal2013making}, LBIQ \cite{tang2011learning} and the current CNN-based state-of-art \cite{kang2014convolutional}.

We perform all our NR-IQA experiments on two widely used datasets - 1. LIVE \cite{sheikh2005live} containing 29 reference images, 779 distorted images, 5 distortion types and 5-7 distortion levels. The images are gray-scale. Through subjective evaluation tests, each user has assigned a score to an image according to its perceptual quality. A metric named difference of mean opinion scores (DMOS) $\in [0, 100]$ was then developed, where 0 indicates highest quality image. 2. TID 2013 \cite{ponomarenko2015image} has 25 reference RGB images, 3000 distorted images, 24 distortion types and 5 distortion levels. In subjective tests, each user was told to assign a score between 0-9 where 9 indicates best visual quality. Mean opinion scores (MOS) were calculated and were provided as ground truth scores/labels for each image. Four of the 24 distortions are common to LIVE and TID datasets. LIVE has all uniform distortions whereas 12 distortions out of total 24 in TID 2013 are non-uniform. See Fig. 1 for an example of uniform/non-uniform distortion. The aim is to learn a mapping between the images and the scores which maximizes Spearman's (SROCC), Pearson's correlation coefficient (PLCC) and Kendall's correlation as those are the standard metrics used for IQA. Since the number of images are so small, we run ours and competing algorithms for 100 splits to remove any data-split bias in all four experiments. As a widely followed convention in IQA, we use $60\%$ of reference and distorted images for training, $20\%$ each for validation and testing everywhere.

The subjective tests conducted for these datasets are extensive. Extreme care was taken to make them statistically unbiased and it is non-trivial to reproduce them. LIVE data creation needed more than $25$K human comparisons for a total of $779$ images. TID 2013 data creation had over 1M visual quality evaluations and $524,340$ comparisons. To avoid geographical bias, the participants came from five countries. In summary, the constraint of small data in such applications comes from the real-world and it is difficult to generate new data or conduct additional tests. The various experiments we perform to evaluate our approach are as follows.

In all the experiments, before we feed training images to the first stage CNN, we preprocess it following the approach of \cite{mittal2012no} that performs local contrast normalization as follows.\vspace{-3pt}

\begin{equation}
\begin{aligned}
&\hat{I}(i,j) = \frac{I(i,j)-\mu(i,j)}{\sigma(i,j)+\epsilon}, \ \mu(i,j) = \sum_{k, l=-3}^3 w_{k,l} I_{k,l} (i,j), \\
&\text{and} \ \sigma(i,j) = \sqrt{\sum_{k,l=-3}^3 w_{k,l} \left(I_{k,l}(i,j) - \mu(i,j)\right)^2}.
\end{aligned}
\end{equation}

\noindent Here, $w$ is a $2$-$D$ circular symmetric Gaussian with 3 standard deviations and normalized to unit volume.

\textbf{Data generation:} The data preparation method of \cite{kang2014convolutional} is common for both datasets. It extracts overlapping $32 \times 32$ patches and then assigns them equal score as that of the image. This strategy works well for \cite{kang2014convolutional} as they only handle LIVE and specific TID distortions that are shared by LIVE. However, to handle non-uniform distortions, we make a small but important modification to their method. We compare the \textit{corresponding} patches of the original image and its distorted counterpart with the help of SSIM \cite{wang2004image}. SSIM is used to measure structural similarity between two images and is robust to slight alterations unlike RMSE. We keep all the patches belonging to a distorted image that have low SSIM scores with their original counterparts. This indicates low similarity between patches which could only point to distortion. We select all patches belonging to reference images. Finally, we make the number of reference and distorted patches equal. We call this method as \textit{selective patch training}. We now describe certain protocols common to all our NR-IQA experiments.

\textbf{Training/testing pipeline:} The method of \cite{kang2014convolutional} trains on $32 \times 32$ patches and averages patch-level scores to get a score for an entire image. We train our first stage on $32 \times 32$ patches obtained by our selection method. The second stage is then trained by hyper-images that are formed using the first-stage patch representations. In our first three NR-IQA experiments, we use the same first stage as that of \cite{kang2014convolutional} to be able to clearly assess the impact of adding a second stage. Addition of a second stage entails little overhead as on LIVE (TID) data, one epoch requires 23 (106) and 3 (43) seconds for both stages respectively. Both the stages as well as \cite{kang2014convolutional} uses mean absolute error as the loss function. 

All the networks are trained using SGD with initial learning rate 0.005 and decaying at a rate of $10^{-5}$ with each update. Nesterov momentum of 0.9 was used. The learning rate was reduced by 10 when the validation error reached a plateau. The first stage was trained for 80 epochs and training was terminated if no improvement in validation error was seen for 20 epochs. The second stage was trained for 200 epochs with the termination criterion set at 40 epochs. Implementations were done in Theano on an Nvidia Tesla K40. We now describe individual experiments. \textit{All the architectures used for various experiments are listed in the supplementary.}

\textbf{Experiment 1:} We evaluate ours and the competing approaches on LIVE. The input sizes for both stages are $32 \times 32$ and $24 \times 23 \times 800$ respectively. Even though LIVE contains only uniform distortions, our approach marginally improves over \cite{kang2014convolutional} 100 splits. This could be due to the better representational capacity of our two stage network as all the image patches contain the same kind of distortion. The results obtained for all the approaches are given in Table  \ref{table:results_NR_IQA}.

\textbf{Experiment 2:} Intuitively, our selective patch training should give us a significant boost in case of TID 2013 data. Since only few patches are noisy, assigning same score to all patches will corrupt the feature space during training. To verify this, we train on TID 2013 data using the approach of \cite{kang2014convolutional} - with and without our selection strategy. Input sizes for both stages are $32 \times 32 \times 3$ and $23 \times 31 \times 800$. We also evaluate our two-stage network to show its superiority over both these approaches. We find that selective patch training boosts Spearman (SROCC) by 0.0992 and Pearson correlation coefficient (PLCC) by 0.072. Two-stage training further improves SROCC by $0.0265$ and PLCC by $0.0111$. The results are given in Table \ref{table:results_NR_IQA}. \textit{Hereinafter, we compare with \cite{kang2014convolutional} assisted with our selective patch training to understand the benefits of the second stage.}

\textbf{Experiment 3:} First, we take four distortions from TID that are shared with LIVE (thus these are uniform). We observe a marginal improvement here for similar reasons as the first experiment. The second part is designed to clearly show the adverse effects of non-uniform, localized distortions on the correlations. Out of 24 distortions, we already have four common ones. We add just two most non-uniform distortions - 1. Non eccentricity pattern noise and 2. Local block-wise distortions. On these six distortions, our approach significantly outperforms that of \cite{kang2014convolutional} with patch selection. Thus to characterize non-uniform distortions, one needs to weight every patch differently, which is exactly what our second stage achieves. Finally, in the third part, we test on the entire TID 2013. To the best of our knowledge, no other learning-based approach has attempted the entire data. The only approach we are aware of that tested on TID is CORNIA \cite{ye2012unsupervised,ye2013real}. However, even they skip two kinds of block distortions. The reasons could be lack of training samples or severe degradation in performance as observed here. We compare our approach with \cite{kang2014convolutional} plus selective patch training. The detailed results are listed in Table \ref{table:results_NR_IQA}. The architecture used was identical to that used in the second experiment.

\textbf{Experiment 4:} We verify that training networks end-to-end from scratch gives poor performance with such low amounts of training data. We define a shallow and a deep CNN of 8 and 14 layers respectively and train them end-to-end on $384 \times 512$ images from TID 2013. Out of all the experiments, this produces the worst performance, making it clear that end-to-end training on such small data is not an option. See Table \ref{table:results_NR_IQA} for results.

\begin{table}[!t]
\centering
\caption{Results of NR-IQA experiments}
\label{table:results_NR_IQA}
\vspace{3pt}
\setlength{\tabcolsep}{0.6em}
\begin{tabular}{c c c}
\multicolumn{3}{c}{\textit{Experiment 1 on LIVE data - 100 Splits}} \\
\hline \\[-9pt]
Approach & SROCC & PLCC \\
\hline \\[-9pt]
DIIVINE & $0.916$ & $0.917$ \\
BLIINDS-II & $0.9306$ & $0.9302$ \\
BRISQUE & $0.9395$ & $0.9424$ \\
CORNIA & $0.942$ & $0.935$ \\
\specialcell{CNN + Selective patch training} & $0.956$ & $0.953$ \\
\textbf{Proposed} & \bm{$0.9581$} & \bm{$0.9585$} \\

\hline \hline \\[-9pt]

\multicolumn{3}{c}{\textit{Experiment 2 on TID data - 100 Splits}} \\
\hline \\[-9pt]
CNN & $0.6618$ & $0.6907$ \\
CNN + Selective patch training & $0.761$ & $0.7627$ \\
\specialcell{\textbf{CNN + Selective patch training} \\ \textbf{+ Stage 2 CNN (proposed)}} & \bm{$0.7875$} & \bm{$0.7738$} \\
\end{tabular}
\begin{tabular}{c c c}
\hline \hline \\[-9pt]
\multicolumn{3}{c}{\textit{Experiment 3 on select distortions of TID - 100 splits}} \\
\hline \\[-9pt]
$\#$ distortions & \specialcell{CNN + Selective \\ patch training \\ (SROCC, PLCC)} & \specialcell{\textbf{Proposed} \\ \textbf{(SROCC,PLCC)}} \\
\hline \\[-9pt]
Four & $0.92$,$0.921$ & \bm{$0.932$},\bm{$0.932$} \\
Six & $0.625$,$0.653$ & \bm{$0.76$},\bm{$0.755$} \\
All (24) & $0.761$,$0.763$ & \bm{$0.788$},\bm{$0.774$} \\
\end{tabular}
\begin{tabular}{c c c}
\hline \hline \\[-9pt]
\multicolumn{3}{c}{\textit{Experiment 4 on TID data - 10 splits}} \\
\hline \\[-9pt]
Approach & SROCC & PLCC \\
\hline \\[-9pt]
Shallow end-to-end CNN & $0.2392$ & $0.4082$ \\
\textbf{Deep end-to-end CNN} & \bm{$0.3952$} & \bm{$0.52$} \\
\hline \hline \\[-9pt]
\multicolumn{3}{c}{\textit{Experiment 5 on TID using pre-trained VGG - 10 splits}} \\
\hline \\[-9pt]
VGG + patch-averaging & $0.6236$ & $0.6843$ \\
\textbf{VGG + second stage CNN} & \bm{$0.6878$} & \bm{$0.713$} \\[1pt]
\hline
\end{tabular}
\end{table}

\textbf{Experiment 5:} A popular alternative when the training data is scarce is to fine-tune a pre-trained network. We took VGG-16, pre-trained on ILSVRC 2014. We used it as the first stage to get patch representations. VGG-16 takes $224 \times 224$ RGB images whereas we have $32 \times 32$ RGB patches. Thus we only consider layers till ``conv$4\_3$'' and get its ReLU-activated responses. All the layers till ``conv$4\_3$'' reduce a patch to a size of $2 \times 2 \times 512$. We append two dense layers of $800$ neurons each and train them from scratch. Rest of the layers are frozen. Please refer to the Caffe VGG prototxt for further architectural details. To train this network, we use a batch size of 256 and a learning rate of 0.01. We average the patch scores obtained from fine-tuned VGG and compute the correlations over $5$ splits. In principle, we should get a performance boost by appending a second stage after VGG, since it would pool in VGG features for all patches and regress them jointly. We use a second stage CNN identical to the one used in experiment 2. We observe that SROCC and PLCC improves by $0.06$ and $0.0287$ respectively. For detailed results, see Table \ref{table:results_NR_IQA}. On the other hand, we see sharp drop in performance for VGG despite it being deep and pre-trained on ImageNet. The reasons for this could be two-fold. As also observed in \cite{kang2014convolutional}, the filters learned on NR-IQA datasets turn out to be quite different than those learned on ImageNet. Thus the semantic concepts represented by the deeper convolutional layers of pre-trained VGG may not be relevant for NR-IQA. Secondly, VGG performs a simple mean subtraction on input images versus we do local contrast normalization (LCN). The latter helps in enhancing the discontinuities (e.g. edges, noise etc.) and suppresses smooth regions, making LCN more suitable for NR-IQA.

Our extensive evaluations on NR-IQA shows that our approach is better at characterizing local distortions present in an image. It improves on the current state-of-art \cite{kang2014convolutional} and other approaches, such as training a shallow/deep network from scratch, or fine-tuning a pre-trained network.

\textbf{Image forgery classification:} In today's age of social media, fake multimedia has become an issue of extreme importance. To combat this, it is necessary to improve detection systems to categorize fake posts. Here, we focus on image forgery/tampering, which is defined as altering an image by various means and then applying post-processing (e.g. blurring) to conceal the effects of forging. Image tampering comes in various forms, for example, copy-move forgery \cite{bayram2009efficient,Sutthiwan2011} and manipulating JPEG headers \cite{farid2009exposing,he2006detecting}. Some other techniques have also been developed to detect forgeries from inconsistent lighting, shadows \cite{fan20123d,kee2010exposing} etc. See the surveys for more information \cite{bayram2008survey,farid2009image}. However, the problem of tampering detection from a single image without any additional information is still eluding researchers. The current state-of-art uses a block-based approach \cite{Sutthiwan2011} which uses block-DCT features. It forms a Markov transition matrix from these features and finally feeds them into a linear SVM. They carry out their experiments on the CASIA-2 tampered image detection database \footnote{\url{http://forensics.idealtest.org/casiav2/}}. It contains 7491 authentic and 5123 tampered images of varying sizes as well as types.

\begin{figure}[!t]
\centering
\includegraphics[width=\linewidth]{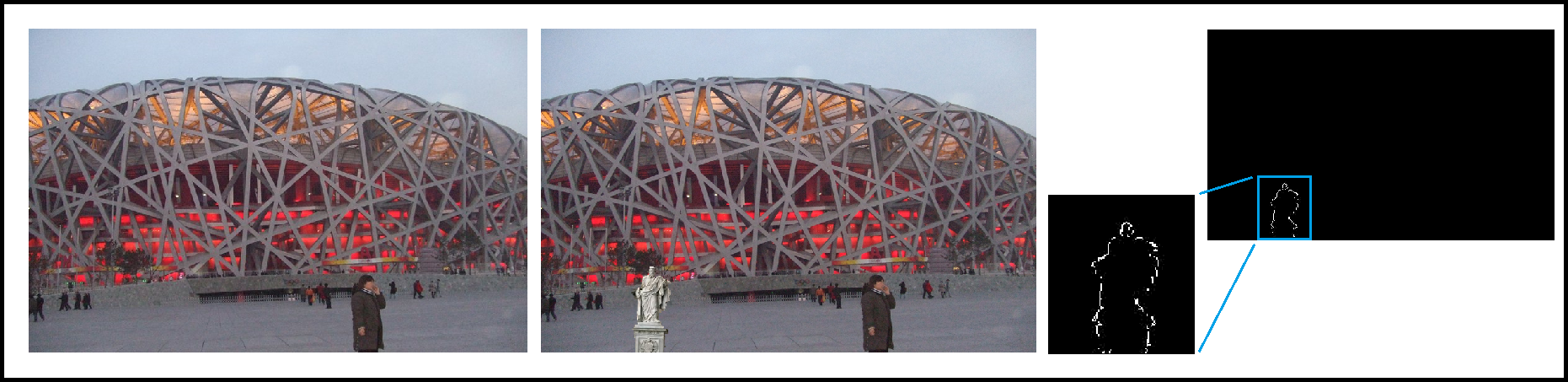}
\caption{Authentic (left) and tampered (middle) image. The resultant contour of the tampered region (right). Please zoom-in and view in color.}
\label{fig:tamperedContour}
\centering
\end{figure}

\textbf{Data generation:} Given a database of authentic and (corresponding) tampered images, we first focus on getting the contour of tampered region(s) by doing image subtraction followed by basic morphological operations. The resultant contour is shown in Fig. \ref{fig:tamperedContour}. We sample 15 pixels along this contour and crop $64 \times 64$ patches by keeping the sampled points as the patch-centroids. Similar to \cite{Sutthiwan2011}, we train on $\frac{2}{3}^{rd}$ of the data and we use $\frac{1}{6}^{th}$ data each for validation and testing. We only subtract the mean of training patches from each patch and do on-the-fly data augmentation by flipping the images. Instead of categorizing \textit{patches} as authentic/tampered, we develop a ranking-based formulation, where the rank of an authentic patch is higher than its counterpart. Note that during testing, we are only given a single image to be classified as authentic or forged and thus contour of forged region cannot be found (or used).

\textbf{Experiment 1:} We train an end-to-end deep CNN that takes an entire image as input and categorizes it as authentic or tampered. The architecture used is given in supplementary. It takes $256 \times 384$ RGB images as input. This size is chosen since it needs minimum number of resize operations over the dataset. The classification accuracy is shown in Table \ref{table:results_forgery}.

\begin{table}[!t]
\centering
\caption{Results of image forgery classification}
\label{table:results_forgery}
\vspace{4pt}
\setlength{\tabcolsep}{0.5em}
\begin{tabular}{c c}
\hline \\[-9pt]
Approach & Classification accuracy \\[2pt]
\hline \hline \\[-9pt]
End-to-end CNN & $75.22\%$ \\
\specialcell{Current State-of-art \\ \cite{Sutthiwan2011}} & $79.20\%$ \\
\textbf{Proposed two stage CNN} & \bm{$83.11\%$} \\
\hline

\end{tabular}
\end{table}

\begin{figure}[!t]
\centering
\includegraphics[width=\linewidth, height=4.0cm]{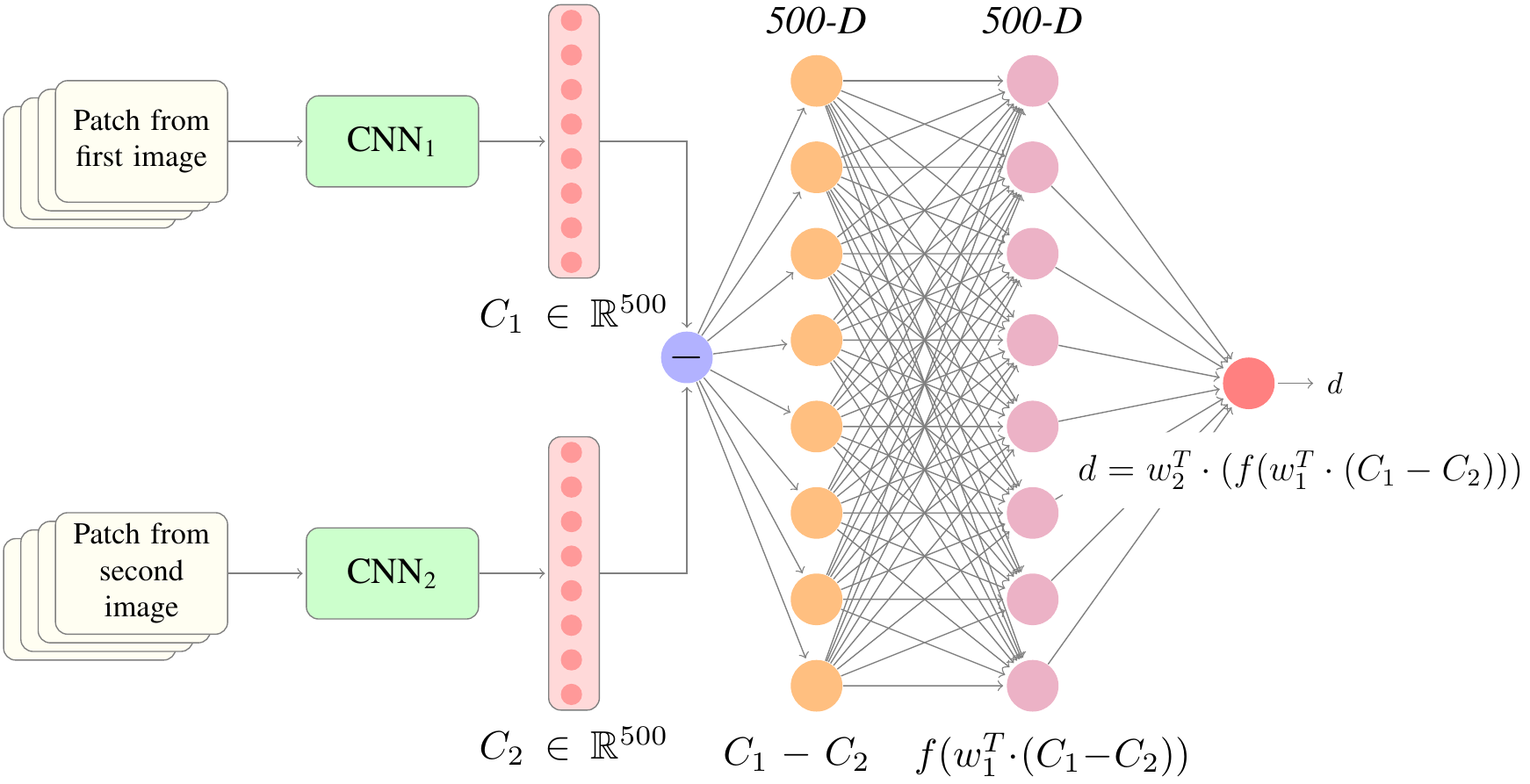}
\caption{Proposed channel architecture. Weight sharing occurs between both channels. Please zoom in to see details.}
\label{fig:tamperingFirstStageArch}
\centering
\end{figure}

\textbf{Experiment 2:} Our first stage CNN learns to rank authentic patches higher than tampered ones. We propose ranking because every patch may contain different amount of forged region or blur. This CNN has two identical channels that share weights. Its architecture is given in supplementary. Let us denote the last dense layer features obtained from an authentic and a tampered patch by $C_{Au}$ and $C_{Tp}$ respectively. We need to learn a weight vector such that $w^T C_{Au} - w^T C_{Tp} > 0$. However, the network trains poorly if we keep feeding the first channel with authentic patches and the second with the tampered ones. Shuffling of patches is necessary in order to achieve convergence. We assign a label of 1 if two channels get authentic and tampered patches (in that order), else -1. Thus we need $d(C_1, C_2) = w_2^T y \cdot (f(w_1^T\cdot(C_1 - C_2)) > 0$, where $C_i$ is the feature from the $i^\text{th}$ channel, whereas $y \in \{-1,1\}$ denotes the label. The transformations achieved through two dense layers and an ReLU are denoted by $w_2(\cdot), w_1(\cdot)$ and $f(\cdot)$ respectively, as shown in Fig. \ref{fig:tamperingFirstStageArch}. The loss function becomes, $L = \max(0,\delta-y \cdot d(C_{1},C_2))$. The term $\max(0,\cdot)$ is necessary to ensure that only non-negative loss gets back-propagated. The $\delta (=3)$ is a user-defined parameter that avoids trivial solutions and introduces class separation.

Stage 1 representation should discriminate between neighborhood patterns along a tampered and an authentic edge (since we chose patches centered on the contour of tampered region). Given an image, we extract patches and form the hyper-image required to train the second stage. We use binary labels, where $1$ denotes authentic image and vice-versa along with a binary cross-entropy loss. Architecture of the second stage is provided in supplementary. To overcome class imbalance, we weigh the samples accordingly. We compare our approach with an end-to-end CNN network (experiment 1) and the current state-of-art in passive, generic image forgery classification \cite{Sutthiwan2011}. CNN-baseline gives the worst performance followed by the current state-of-art. This is expected since the latter extracts \textit{block-level} DCT features whereas the CNN-baseline tries to learn from an entire image - a considerably difficult task especially when the tampered/forged parts are localized and well camouflaged. Our hybrid approach beats the CNN-baseline and the state-of-art by $8\%$ and $4\%$ respectively. All these experiments underlines the importance of collectively learning from image patches when the data is scarce and shows flexibility of our approach.

\section{Conclusion} \label{sec:conclusion}

We presented the notion of CNN-based hyper-image representations. Our training scheme involving these hyper-images excels in scenarios where the label is dependent on the localized artifacts in an image. In these networks, the first stage is only responsible for learning the discriminative representations of small image patches. The second stage collectively considers all the patches of an image unlike many other previous approaches. It optimally weighs and pools all the patches, and develops a mapping between them and the image label. Our approach enables us to train networks with greater representational capacity than their conventional counterparts. We observe in all our experiments that the second stage always provides us with a significant improvement. We apply our approach to a synthetic and two challenging vision tasks - NR-IQA and image forgery classification. Our approach comfortably outperforms other CNN-baselines as well as the existing state-of-art approaches.




%


\begin{table*}[p]
\centering
\caption{Architectures of deep networks used in this paper. The term $C(n, r, s)$ denotes $r \times r$, $n$ ``same'' convolutional filters with stride $s$. We omit $r$ and $s$ when $r = 3$ and $s = 1$. MP$(n)$ is a max-pooling reduce the image size by a factor of $n$. FC(n) and Drop(n) denote dense layer with $n$ neurons and dropout rate of $n$ respectively.}
\label{table:deepArchitectures}
\vspace{3pt}
\setlength{\tabcolsep}{0.42em}
\begin{tabular}{c c}
\hline \hline \\[-9pt]
Architecture & Layer descriptions \\
\hline \\[-9pt]

\textbf{Synthetic stage 1} & \specialcell{Input($128,128$) -- C(16) -- MP(2) -- C(32) -- MP(2) -- 2 $\times$ C(48) -- MP(2) --  2 $\times$ C(64) -- \\ MP(2) -- 2 $\times$ C(128) -- MP(2) -- FC(400) -- Drop(0.5) -- FC(400) -- Drop(0.5) -- FC(1,`linear')} \\[8pt]

\textbf{Synthetic stage 2} & \specialcell{Input(10,10,400) -- 2 $\times$ C(16) -- MP(2) -- 2 $\times$ C(32) -- MP(2) -- 2 $\times$ C(64) --  \\ MP(2) -- FC(400) -- Drop(0.5) -- FC(400,`tanh') -- Drop(0.5) -- FC(1,`linear')} \\[8pt]

\textbf{LIVE/TID stage 1} & Please refer to Table \ref{table:stage1} on the next page \\[8pt]

\textbf{LIVE stage 2} & \specialcell{Input(24,23,800) -- 2 $\times$ C(32) -- MP(2) -- 2 $\times$ C(48) -- MP(2) -- 2 $\times$ C(64) -- MP(2) -- \\2 $\times$ C(128) -- MP(2) -- FC(500) -- Drop(0.5) -- FC(500,`tanh') -- Drop(0.5) -- FC(1,`linear')} \\[8pt]

\textbf{TID stage 2} & \specialcell{Input(23,31,800) -- 2 $\times$ C(64) -- MP(2) -- 2 $\times$ C(64) -- MP(2) -- 2 $\times$ C(128) -- MP(2) -- \\2 $\times$ C(128) -- MP(2) -- FC(500) -- Drop(0.5) -- FC(500,`tanh') -- Drop(0.5) -- FC(1,`linear')} \\[8pt]

\textbf{\specialcell{Shallow end-to-end \\network for TID (Expt 4)}} & \specialcell{Input(384, 512, 3) -- C(32, 7, 2) -- MP(2) -- C(64, 7, 1) -- MP(2) -- 2 * ( C(128) -- MP(2) ) \\-- 2 * ( C(256) ) -- MP(2) -- FC(400) -- Drop(0.25) -- FC(400, 'tanh') -- Drop(0.25) -- FC(1,'linear')} \\[8pt]

\textbf{\specialcell{Deep end-to-end network \\for TID (Expt 4)}} & \specialcell{Input(384, 512, 3) -- C(32) -- MP(2) -- C(64) -- MP(2) -- 2 * ( C(128) -- MP(2) ) - \\2 * ( C(256) - C(256) - MP(2) ) -- 2 * (C(512) -- C(512) -- MP(2) ) -- \\FC(400) - Drop(0.25) -- FC(400, 'tanh') -- Drop(0.25) -- FC(1,'linear')} \\[8pt]

\textbf{Forgery channel} & \specialcell{Input(64,64,3) -- 2 $\times$ C(64) -- MP(2) -- 2 $\times$ C(128) -- MP(2) -- 2 $\times$ C(128) -- \\MP(2) -- 2 $\times$ C(256) -- MP(2) -- FC(500) -- Drop(0.5) -- FC(500) -- Drop(0.5)} \\[8pt]

\textbf{Forgery stage 2} & \specialcell{Input(15,15,500) -- 3 $\times$ C(64) -- MP(2) -- 3 $\times$ C(128) -- MP(2) -- 3 $\times$ C(256) -- \\MP(2) -- FC(800) -- Drop(0.5) -- FC(800) -- Drop(0.5) -- FC(1,`sigmoid')} \\[8pt]

\specialcell{\textbf{Forgery end-}\\\textbf{to-end CNN}} & \specialcell{Input(256,384,3) -- C(32) -- MP(2) -- C(64) -- MP(2) -- 2 $\times$ C(64) -- MP(2) -- 2 $\times$ C(128) -- MP(2) -- 2 $\times$ \\ C(128) -- MP(2) -- 2 $\times$ C(256) -- MP(2) --FC(500) -- Drop(0.5) -- FC(500) -- Drop(0.5) -- FC(1,`sigmoid')}  \\[3pt]

\hline
\hline

\end{tabular}
\end{table*}



\begin{table*}[p]
\centering
\caption{First stage architecture for LIVE/TID data used in this paper.}
\label{table:stage1}
\vspace{3pt}
\setlength{\tabcolsep}{0.42em}
\begin{tabular}{c c}
\hline \hline \\[-9pt]
Layer Details & Connected to \\
\hline \\[-9pt]

Input -- $32 \times 32 \times 3$ (for TID) and $32 \times 32$ (for LIVE) & --\\[8pt]

Convolutional layer -- $50 \times 7 \times 7$, name = 'Conv1' & Input \\[8pt]

\hline \\[-9pt]

Max-pooling: size = (26,26), stride = (26,26) , name = 'MaxPool' & Conv1 \\[8pt]
Min-pooling: size = (26,26), stride = (26,26), name = 'MinPool' & Conv1 \\[8pt]

\hline \\[-9pt]

Concatenation layer: inputs = (MinPool, MaxPool), name = 'Concat' & -- \\[8pt]
Fully-connected layer: 800 nodes, name = 'FC1' & Concat \\[8pt]
Fully-connected layer: 800 nodes, name = 'FC2' & FC1 \\[8pt]
Output node, name = 'output' & FC2 \\[8pt]

\hline \\[-9pt]

Loss = 'Mean absolute error' & -- \\[8pt]

\hline
\hline

\end{tabular}
\end{table*}



{\small
\bibliographystyle{aaai}
\bibliography{egbib}
}

\end{document}